\documentclass[conference, a4paper, pdftex]{IEEEtran}
\IEEEoverridecommandlockouts
\usepackage{cite}
\usepackage{amsmath,amssymb,amsfonts}
\usepackage{algorithmic}
\usepackage{graphicx}
\usepackage{textcomp}
\usepackage{xcolor}
\makeatletter
\newcommand{\linebreakand}{%
  \end{@IEEEauthorhalign}
  \hfill\mbox{}\par
  \mbox{}\hfill\begin{@IEEEauthorhalign}
}
\makeatother
\def\BibTeX{{\rm B\kern-.05em{\sc i\kern-.025em b}\kern-.08em
    T\kern-.1667em\lower.7ex\hbox{E}\kern-.125emX}}
\begin{document}
\title{Classification in Japanese Sign Language \\Based on Dynamic Facial Expressions}

\author{\IEEEauthorblockN{Yui Tatsumi}
\IEEEauthorblockA{\textit{School of FSE,} \\
\textit{Waseda University}\\
Tokyo, Japan}
\and
\IEEEauthorblockN{Shoko Tanaka}
\IEEEauthorblockA{\textit{School of FSE,} \\
\textit{Waseda University}\\
Tokyo, Japan}
\and
\IEEEauthorblockN{Shunsuke Akamatsu}
\IEEEauthorblockA{\textit{Graduate School of FSE,} \\
\textit{Waseda University}\\
Tokyo, Japan}
\linebreakand
\IEEEauthorblockN{Takahiro Shindo}
\IEEEauthorblockA{\textit{Graduate School of FSE,} \\
\textit{Waseda University}\\
Tokyo, Japan}
\and
\IEEEauthorblockN{Hiroshi Watanabe}
\IEEEauthorblockA{\textit{Graduate School of FSE,} \\
\textit{Waseda University}\\
Tokyo, Japan}
}

\maketitle

\begin{abstract}
Sign language is a visual language expressed through hand movements and non-manual markers. Non-manual markers include facial expressions and head movements. These expressions vary across different nations.
Therefore, specialized analysis methods for each sign language are necessary.
However, research on Japanese Sign Language (JSL) recognition is limited due to a lack of datasets.
The development of recognition models that consider both manual and non-manual features of JSL is crucial for precise and smooth communication with deaf individuals.
In JSL, sentence types such as affirmative statements and questions are distinguished by facial expressions.
In this paper, we propose a JSL recognition method that focuses on facial expressions.
Our proposed method utilizes a neural network to analyze facial features and classify sentence types.
Through the experiments, we confirm our method's effectiveness by achieving a classification accuracy of 96.05\%.

\end{abstract}
\section{Introduction}
In Japan, communication methods that rely on knowledge of the Japanese language are frequently used between deaf and hearing individuals.
For instance, there are tools such as written communication and speech-to-text applications.
However, many deaf individuals struggle with communicating in Japanese because Japanese Sign Language (JSL) has its unique vocabulary and grammar, separate from Japanese. 
Furthermore, many hearing individuals are not familiar with JSL. 
The development of JSL recognition methods is required in order to ensure precise and smooth communication between deaf and hearing.

Previous research on sign language recognition has primarily focused on American, German, and Chinese sign languages. 
Moreover, these studies often concentrate on hand movements and apply hand pose estimation techniques for recognition.
However, non-manual markers such as facial expressions and body orientation are indispensable for a complete understanding of sign language sentences.

In this paper, we propose a recognition method for JSL that focuses on non-manual markers. 
These markers have significant impact on syntactic and semantic information. 
For example, when hand gestures expressed in an affirmative statement are accompanied by facial expressions such as wide-open eyes, raised eyebrows, and a tucked chin, the sentence transforms into a Yes/No-question. 
When paired with repeated weak head shakes and furrowed eyebrows, it is classified as a WH-question.
Using a neural network, we analyze facial expressions to distinguish the affirmative sentence, Yes/No-question, and WH-question.

\section{Related Work}
\subsection{Datasets}\label{DS}
In sign language recognition research, extensive publicly available datasets for American, German, and Chinese sign languages are widely used \cite{b1}. These datasets include videos of individuals using sign language, as well as the corresponding translations. However, datasets for JSL are limited in number; as a result, there are few studies on the topic. To address this issue, in this paper, we create JSL video datasets and apply effective data augmentation.
\subsection{Methods}\label{M}
Numerous previous studies on sign language recognition utilize hand poses. Wang \textit{et al.} combine object detection and hand pose estimation to detect hand shapes and recognize the alphabet and numerals in American Sign Language (ASL) \cite{b2}. Similarly, Chu \textit{et al.} and Wu \textit{et al.} focus on hand poses to recognize Japanese and Chinese Sign Language, respectively \cite{b3}, \cite{b4}.
However, non-manual markers such as facial expressions are also important in sign language recognition and should be focused on. A study on ASL proposes a recognition method employing facial expressions analysis \cite{b5}. In their study, the exact location of facial feature points is reconstructed by correcting the tracked points based on learned face shape subspaces. 
The extracted data are then analyzed by a recognition system to identify six non-manual markers in ASL.

\section{Proposed Method}
\begin{figure}[t]
  \centering
\includegraphics[width=\hsize, height=4.5cm]{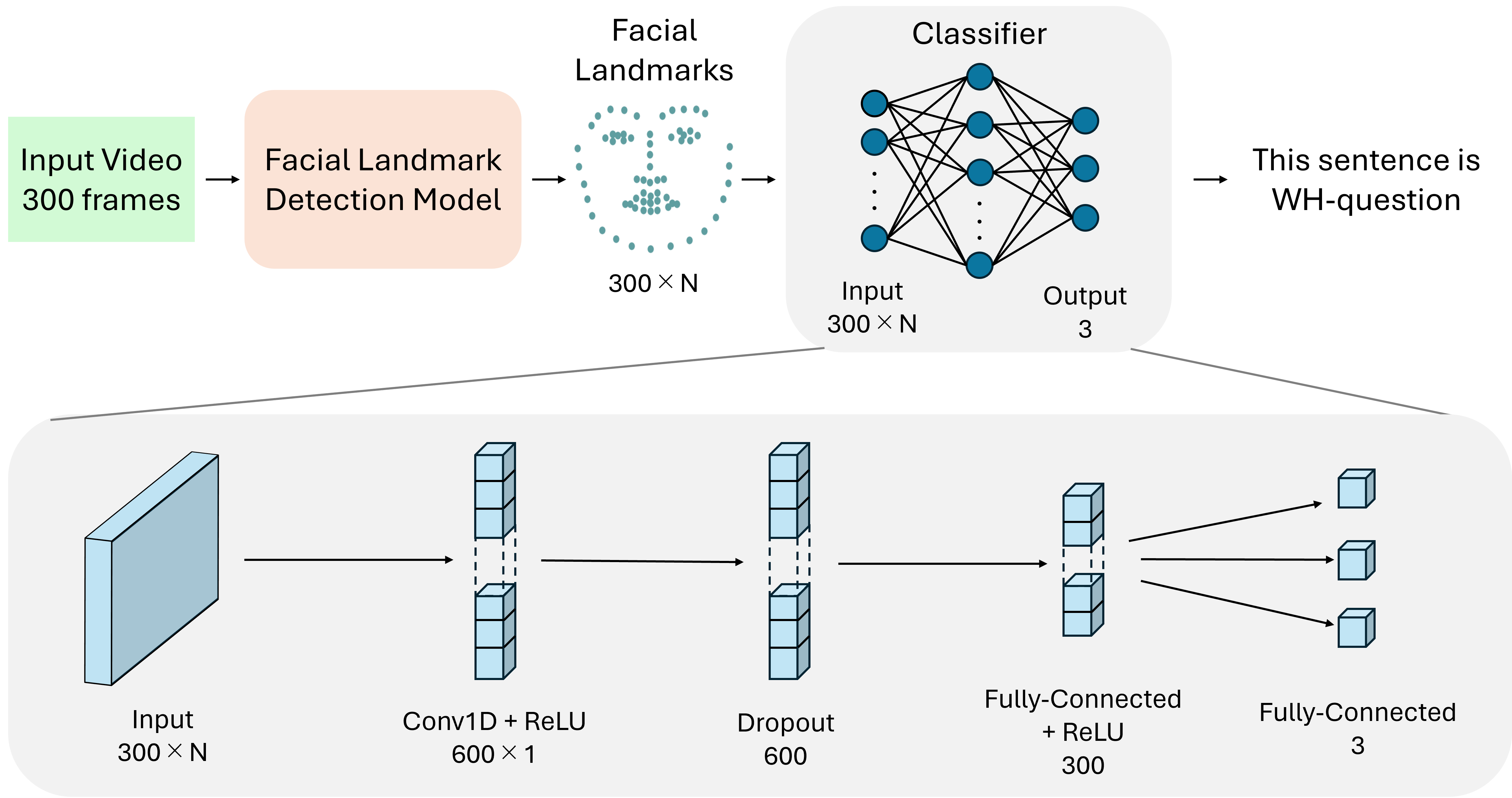}
  \caption{An overview of our proposed method. The variable N indicates the number of facial landmarks extracted from OpenPose, MediaPipe, and Dlib models, which are 140, 936, and 136, respectively.}
  \label{fig:Overview}
\end{figure}

We propose a JSL recognition method focusing on its facial expressions. The overview of our proposed method is shown in Fig. \ref{fig:Overview}. This method classifies JSL videos into three classes: (1) affirmative sentences, (2) Yes/No questions, or (3) WH-questions, based on the facial expressions of the signers. Initially, sign language videos are input into a model for detecting facial landmarks. In our experiments, OpenPose \cite{b6}, MediaPipe \cite{b7}, and Dlib \cite{b8} are utilized as the model for comparison. The outputs from the models include $x$ and $y$ coordinates of 70, 468, and 68 estimated facial landmarks, respectively. Then, these landmarks are utilized by a neural network based classifier to classify JSL sentences. The classifier consists of a convolutional layer, an activation function (ReLU), and two fully-connected layers. Cross Entropy Loss is used as the loss function.

\section{Experiment}
\subsection{Datasets}
A total of 378 JSL videos are collected as our dataset. Among these, 12 videos are created by Morita \textit{et al.} \cite{b9}, and 39 videos by Oka \textit{et al.} \cite{b10}. These videos are performed by three native JSL signers who are deaf. The remaining 327 videos are created by us. The signers are four individuals from Waseda University Sign Language Club, including a deaf student. Each of the 378 videos captures a single participant expressing a brief sentence in JSL within 10 seconds. The frame rate of the videos is 30 fps. Each sentence is labeled according to three categories: affirmative, Yes/No-question, and WH-question. 302 videos are used for training and 76 videos are employed for validation.

\subsection{Training and Validation}
For training, initially, each video is uniformly extended to 300 frames by padding with the value from its final frame. Secondly, facial landmarks are extracted using the facial landmark detection model. These landmarks are normalized so that the average distance from the nose tip to the other coordinates is one. In addition, permutation is applied as a data augmentation strategy \cite{b11}. Finally, the classifier is trained using normalized data of 302 videos.

For validation, 76 videos are employed. The validation metrics include accuracy, precision, recall, and F1 score. Table \ref{table:results} shows the results.
The classification accuracy achieved using OpenPose demonstrates the validity of our dataset and data augmentation, as well as the effectiveness of our proposed method.
Additionally, employing OpenPose as the detection model results in higher accuracy compared to both MediaPipe and Dlib. This superiority stems from OpenPose's robust ability to detect facial landmarks in videos with cluttered backgrounds and sudden movements, such as head shakes.

\begin{table}[t]
  \begin{center}
  \caption{Classification Results Using Various Detection Models}
  \label{table:results}
  \begingroup
  \setlength{\tabcolsep}{2.8pt}
    \begin{tabular}{c|cccc} \hline
      \begin{tabular}{c}
      Facial Landmark \\ Detection Model
      \end{tabular}
      & Accuracy(\%) & Precision(\%) & Recall(\%)  & F1 Score(\%) \\ \hline \hline
       OpenPose \cite{b6}& \textbf{96.05}  & \textbf{96.25}  & \textbf{96.05} & \textbf{96.12} \\ 
       MediaPipe \cite{b7}& 88.16 & 88.48 & 88.25 & 88.34 \\
       Dlib \cite{b8}& 82.89 & 83.59 & 82.91 & 83.10 \\ \hline
    \end{tabular}
    \endgroup
  \end{center}
\end{table}

\section{Conclusion}
In this paper, we propose a JSL recognition technique that focuses on its facial expressions. We construct a classifier that categorizes JSL videos into affirmative statements, Yes/No-questions, or WH-questions based on various facial features. Experimental results present the effectiveness of our proposed method. Future work is required to achieve comprehensive recognition of JSL by combining hand pose estimation techniques.

\end{document}